\documentclass{article}
\usepackage{spconf,amsmath,graphicx,hyperref}
\usepackage{cite}
\usepackage{pgfplots}
\pgfplotsset{compat=1.18}
\usepackage{booktabs}
\usepackage{multirow}
\usepackage{booktabs}
\usepackage{hyperref}

\title{THE IMPACT OF AUDIO WATERMARKING ON AUDIO ANTI-SPOOFING COUNTERMEASURES}
\name{ZhenShan Zhang$^{1}$, Xueping Zhang$^{1}$, Yechen Wang$^{2}$, Liwei Jin$^{2}$, Ming Li$^{1}$  \thanks{Corresponding Author: Ming Li, ming.li369@dukekunshan.edu.cn}}

\address{$^{1}$Suzhou Municipal Key Laboratory of Multimodal Intelligent Systems,\\ Digital Innovation Research Center,
Duke Kunshan University, Kunshan, China \\ $^{2}$OfSpectrum, Inc., Los Angeles, USA}
\begin{document}
%
\maketitle
\begin{abstract}
This paper presents the first study on the impact of audio watermarking on spoofing countermeasures. While anti-spoofing systems are essential for securing speech-based applications, the influence of widely used audio watermarking—originally designed for copyright protection—remains largely unexplored. We construct watermark-augmented training and evaluation datasets, named the \textit{Watermark-Spoofing dataset}, by applying diverse handcrafted and neural watermarking methods to existing anti-spoofing datasets. Experiments show that watermarking consistently degrades anti-spoofing performance, with higher watermark density correlating with higher Equal Error Rates (EERs). To mitigate this, we propose the \textit{Knowledge-Preserving Watermark Learning (KPWL)} framework, enabling models to adapt to watermark-induced shifts while preserving their onriginal-domain spoofing detection capability. These findings reveal audio watermarking as a previously overlooked domain shift and establish the first benchmark for developing watermark-resilient antispoofing systems.
All related protocols are publicly available at \url{https://github.com/Alphawarheads/Watermark_Spoofing.git}
.

\end{abstract}
\section{Introduction}
\label{sec:intro}

In recent years, advances in speech synthesis and voice conversion have enabled highly realistic artificial speech, raising serious concerns about security in voice-based applications. Automatic speaker verification (ASV) and its anti-spoofing countermeasures have thus become essential for safeguarding applications such as banking, forensics, and access control \cite{Andre2021Manipu}. Meanwhile, audio watermarking is widely used for copyright protection and content authentication \cite{HUA2016222}.

Most watermarking research has focused on imperceptibility and robustness to common signal processing operations \cite{PAVLOVIC2022103381,chen2023wavmark,timbrewatermarking-ndss2024}, while antispoofing has advanced with self-supervised front ends like XLS-R\cite{lv2022fakeaudio} and WavLM\cite{chen2022wavlm}, combined with strong back-end models such as AASIST\cite{jung2022aasist}, SLS\cite{zhang2024xlsrsls}, and Nes2Net\cite{liu2025nes2net}. However, despite growing research on adversarial perturbations, the impact of audio watermarking—introducing structured yet persistent perturbations—on antispoofing systems remains unexplored.

 
Although audio watermarks are designed to be imperceptible to human listeners, they inevitably modify the underlying speech signal distribution~\cite{HUA2016222,alsabhany2020digital}, especially in real-world scenarios where multiple watermarking schemes may coexist. Such modifications can introduce complex and previously unstudied domain shifts between original and watermarked audio, potentially leading spoofing detectors to misclassify genuine or synthetic speech. This issue is critical for cross-dataset generalization, where antispoofing models must handle unseen spoofing techniques\cite{wang2020asvspoof2019}, and unexamined interactions with watermarking could compromise their security benefits.

\noindent\textbf{Our main contributions are as follows:}
\begin{itemize}
\item \textbf{Watermark-Spoofing Dataset:} We create dedicated training and evaluation datasets for anti-spoofing models under watermark-induced domain shifts.

\item \textbf{Extensive evaluation:} A diverse set of handcrafted and DNN-based watermarking methods~\cite{timbrewatermarking-ndss2024,roman2024proactivedetectionvoicecloning,singh24_interspeech,zhao2021fsvc,nugraha2011dsss,dhar2015blindsvd,adhiyaksa2022reversible,mushtaq2024blind} is systematically applied to benchmark datasets~\cite{yamagishi2021asvspoof,wang2020asvspoof2019,muller2022does} to assess their impact on state-of-the-art spoofing detection models.

\item \textbf{Adaptation framework:} We propose the \textit{Knowledge-Preserving Watermark Learning (KPWL)}, a framework that mitigates watermark-induced performance degradation while preserving original-domain detection accuracy.
\end{itemize}


\section{Watermarking Impact Analysis}
\label{sec:typestyle}

In this section, the impact of audio watermarking on spoofing detection is evaluated. A set of six handcrafted watermarks~\cite{dhar2015blindsvd,yang2024improvedphasecoding,alsabhany2020digital,zhao2021fsvc,natgunanathan2017patchwork,mushtaq2024blind} 
and three DNN (Deep Neural Network)-based methods~\cite{chen2023wavmark,timbrewatermarking-ndss2024,roman2024proactivedetectionvoicecloning}  is applied to benchmark datasets and assessed using three state-of-the-art antispoofing models\cite{tak2022automatic,zhang2024xlsrsls,liu2025nes2net}. The Equal Error Rate (EER) of each model is measured as the proportion of watermarked samples in the evaluation data increases. Each watermarked evaluation set maintains a strict 1:1 ratio between DNN-based and handcrafted watermark types, and all watermarking methods within each group are represented in equal proportion. Furthermore, each type of watermark is applied to both bonafide and spoof samples proportionally to their natural distribution within the corresponding dataset. This setup is to simulate the cases that both bonafide and spoofing samples could be protected by different kinds of watermarking algorithms. This balanced design ensures that any observed performance differences can be attributed to the presence of watermarking, reducing the effect of class imbalance or unequal representation of specific watermarking methods.

\subsection{Watermark-Spoofing Dataset}
A mixed watermark dataset comprising six handcrafted methods~\cite{dhar2015blindsvd,yang2024improvedphasecoding,alsabhany2020digital,zhao2021fsvc,natgunanathan2017patchwork,mushtaq2024blind} and three DNN-based methods~\cite{chen2023wavmark,timbrewatermarking-ndss2024,roman2024proactivedetectionvoicecloning} is constructed to approximate real-world conditions.
This collection is applied to three benchmark antispoofing datasets—ASVspoof~2021 LA, ASVspoof~2021 DF (LA21, DF21)~\cite{yamagishi2021asvspoof}, and the In-the-Wild dataset (ITW)~\cite{muller2022does}.
For each dataset, three watermarked variants are created, with 75\%, 50\%, and 25\% of the samples watermarked while the remaining samples are kept clean, following the distribution described above.
Together, these corpora constitute the \textit{Watermark-Spoofing Seen Evaluation dataset}.
Similarly, 50\% of the ASVspoof~2019 LA (LA19) training data~\cite{wang2020asvspoof2019} is watermarked using the same configuration, forming the \textit{Watermark-Spoofing Training dataset}.

\subsection{Impact of Watermark Data on Anti-Spoofing}

\textbf{Anti-Spoofing on Single Watermark Data: }
To quantify the impact of individual watermarking methods, five representative DNN-based watermarks—RobustDNN\cite{PAVLOVIC2022103381}, WavMark\cite{chen2023wavmark}, Timbre~\cite{timbrewatermarking-ndss2024}, and AudioSeal~\cite{roman2024proactivedetectionvoicecloning}—are applied to the In-The-Wild evaluation dataset(ITW)\cite{muller2022does}.
Equal Error Rates (EERs) are measured under varying watermark ratios using a XLSR+SLS \cite{zhang2024xlsrsls} baseline detector trained on ASVspoof~2019(LA19)\cite{wang2020asvspoof2019}.
\begin{table}[t]
\raggedright
\small
\setlength{\tabcolsep}{3pt} 
\caption{EER (\%) on ITW dataset under different watermark ratios. \textit{$\Delta$} shows relative degradation compared to the clean (0\%) setting.}
\label{tab:wm_single}
\begin{tabular}{l|cccc|c}
\toprule
\textbf{Method} & \textbf{75\%} & \textbf{50\%} & \textbf{25\%} & \textbf{0\%} & 75\%$\Delta(\%)$ \\
\midrule
AudioSeal (2024)\cite{roman2024proactivedetectionvoicecloning} & 7.46 & 7.40 & 7.35 & 7.32 & 1.91\% \\
Timbre (2023)\cite{timbrewatermarking-ndss2024}   & 8.18 & 7.93 & 7.53 & 7.32 & 11.75\% \\
WavMark (2023)\cite{chen2023wavmark}   & 9.90 & 9.06 & 8.23 & 7.32 & 35.25\% \\
DNN (2022) \cite{PAVLOVIC2022103381}      & 9.06 & 8.65 & 8.06 & 7.32 & 23.77\% \\
\bottomrule
\end{tabular}
\end{table}

Table \ref{tab:wm_single} shows that single watermarking indeed degrades spoofing detection performance, with EERs increasing as the watermark proportion grows. Moreover, the observed degradation tends to diminish with newer watermarking methods, suggesting that recent designs introduce less disruptive perturbations.

\textbf{Anti-Spoofing on Mixed Watermark Data: }
The \textit{Watermark-Spoofing Seen Evaluation dataset} is then used to assess three state-of-the-art models trained on the LA19 training set: XLSR-AASIST~\cite{tak2022automatic}, XLSR-SLS~\cite{zhang2024xlsrsls}, and XLSR-Nes2Net-X~\cite{liu2025nes2net}.
As shown in Table \ref{tab:eer_mixed}, results across all models and datasets reveal a clear positive correlation: higher proportions of watermarked samples consistently lead to increased EERs, demonstrating that watermarking degrades spoofing detection performance.

\begin{table}[t]
\centering
\small   
\setlength{\tabcolsep}{3pt} 
\caption{EER (\%) of three baseline models trained on LA19 and evaluated on \textit{Watermark-Spoofing Seen Evaluation dataset} sets at different ratios.}
\label{tab:eer_mixed}
\begin{tabular}{c |c| cccc |c}
\toprule
\textbf{Model} & \textbf{Test Data} & \textbf{75\%} & \textbf{50\%} & \textbf{25\%} & \textbf{0\%} & 75\%$\Delta(\%)$ \\
\midrule
\multirow{3}{*}{\shortstack{XLSR\\+AASIST\\(2022)}} 
  & LA21 & 0.88 & 0.83 & 0.79 & 0.73 & 20.55 \\
  & ITW  & 11.28 & 10.65 & 10.00 & 9.42 & 19.75 \\
  & DF21 & 6.16 & 6.08 & 5.99 & 5.86 & 5.12 \\
\midrule
\multirow{3}{*}{\shortstack{XLSR\\+SLS\\(2024)}} 
  & LA21 & 3.68 & 3.52 & 3.35 & 3.02 & 21.85 \\
  & ITW  & 8.46 & 7.83 & 7.57 & 7.32 & 15.57 \\
  & DF21 & 2.23 & 2.13 & 2.17 & 2.01 & 10.94 \\
\midrule
\multirow{3}{*}{\shortstack{XLSR\\+Nes2Net-X\\(2025)}} 
  & LA21 & 2.25 & 2.20 & 2.08 & 2.00 & 12.50 \\
  & ITW  & 6.40 & 6.07 & 5.84 & 5.50 & 16.36 \\
  & DF21 & 1.85 & 1.84 & 1.82 & 1.76 & 5.11 \\
\bottomrule
\end{tabular}
\end{table}

Furthermore, as shown in Table \ref{tab:xlsr_sls_mixed}, a multi-dataset\cite{wang2020asvspoof2019,reimao2019for,salvi2023timittts,yaroshchuk2023opensynth} trained XLSR+SLS model, achieving state-of-the-art results on both experimental(LA21 and DF21) and real-world (ITW) deepfake datasets, also suffers a notable performance drop when tested on watermarked audio. Even the advanced model struggles under watermarked conditions, suggesting that audio watermarking introduces a previously unstudied form of domain shift that current antispoofing systems are not designed to adapt to or be robust against.
\begin{table}[t]
\centering
\small
\setlength{\tabcolsep}{4pt}
\caption{EER (\%) of XLSR+SLS trained on TIMIT\cite{salvi2023timittts}, FoR\cite{reimao2019for}, ODSS\cite{yaroshchuk2023opensynth} and ASV19\cite{wang2020asvspoof2019} and evaluated under 3 datasets with \textit{Watermark-Spoofing Seen Evaluation dataset}. $\Delta$ denotes the relative increase from 0\%.}
\label{tab:xlsr_sls_mixed}
\begin{tabular}{l| cccc |c}
\toprule
\textbf{Dataset} & \textbf{75\%} & \textbf{50\%} & \textbf{25\%} & \textbf{0\%} & 75\%$\Delta(\%)$ \\
\midrule
LA21 & 2.82 & 2.67 & 2.55 & 2.46 & 14.63\% \\
ITW  & 4.06 & 3.77 & 3.49 & 3.17 & 28.08\% \\
DF21 & 1.18 & 1.10 & 1.04 & 1.02 & 15.69\% \\
\bottomrule
\end{tabular}
\end{table}

\textbf{Limitations of Direct Training for Watermarked Anti-Spoofing: }
To further examine the domain shift induced by watermarking, the \textit{Watermark-Spoofing Training dataset} was used to train the three selected models—XLSR-AASIST, XLSR-SLS, and XLSR-Nes2Net-X.

As displayed in Table \ref{tab:eer_train_wm}, the results were complex: models such as XLSR+SLS and XLSR+Nes2Net-X appeared to extract more stable features and partially suppress watermark-induced shifts as the watermark density increased, albeit at the cost of degraded performance on the original clean evaluation sets.

In contrast, XLSR+AASIST showed more dramatic changes. On the In-The-Wild dataset, it exhibited behavior resembling data augmentation, achieving slight performance gains, while on the ASVspoof2021 LA dataset its performance dropped sharply from a state-of-the-art 0.78\% EER to an average level of 3.84\%. This degradation may be linked to a reduction of overfitting on LA21 caused by the added watermark noise.

These observations indicate that introducing watermarked audio into training does not provide a consistent benefit and may distort model learning in unexpected ways. 

\begin{table}[t]
\centering
\small
\setlength{\tabcolsep}{4pt}
\caption{EER (\%) of three models trained with \textit{Watermark-Spoofing Training dataset}, evaluated \textit{Watermark-Spoofing Seen Evaluation dataset}.}
\label{tab:eer_train_wm}
\begin{tabular}{c |c| cccc}
\toprule
\textbf{Model} & \textbf{Dataset} & \textbf{75\%} & \textbf{50\%} & \textbf{25\%} & \textbf{0\%} \\
\midrule
\multirow{3}{*}{\shortstack{XLSR\\+AASIST\\(2022)}} 
  & LA21 & 4.41 & 4.18 & 3.96 & 3.84 \\
  & ITW  & 10.08 & 9.56 & 9.16 & 8.73 \\
  & DF21 & 3.11 & 2.98 & 2.85 & 2.66 \\
\midrule
\multirow{3}{*}{\shortstack{XLSR\\+SLS\\(2024)}} 
  & LA21 & 3.28 & 3.25 & 3.23 & 3.17 \\
  & ITW  & 9.03 & 8.75 & 8.57 & 8.21 \\
  & DF21 & 2.13 & 2.07 & 2.00 & 1.87 \\
\midrule
\multirow{3}{*}{\shortstack{XLSR\\+Nes2Net-X\\(2025)}} 
  & LA21 & 2.89 & 2.83 & 2.79 & 2.75 \\
  & ITW  & 7.58 & 7.23 & 6.90 & 6.58 \\
  & DF21 & 1.92 & 1.90 & 1.88 & 1.87 \\
\bottomrule
\end{tabular}
\end{table}

\section{Methodology}
\label{sec:majhead}
As shown in the previous sections, audio watermarking introduces a notable domain shift that degrades the performance of antispoofing models. While directly incorporating watermarked data into training can sometimes suppress watermark-induced changes, the results are highly unstable and often compromise the model’s performance on clean, unwatermarked data.

\begin{figure}
    \centering
    \includegraphics[width=1\linewidth]{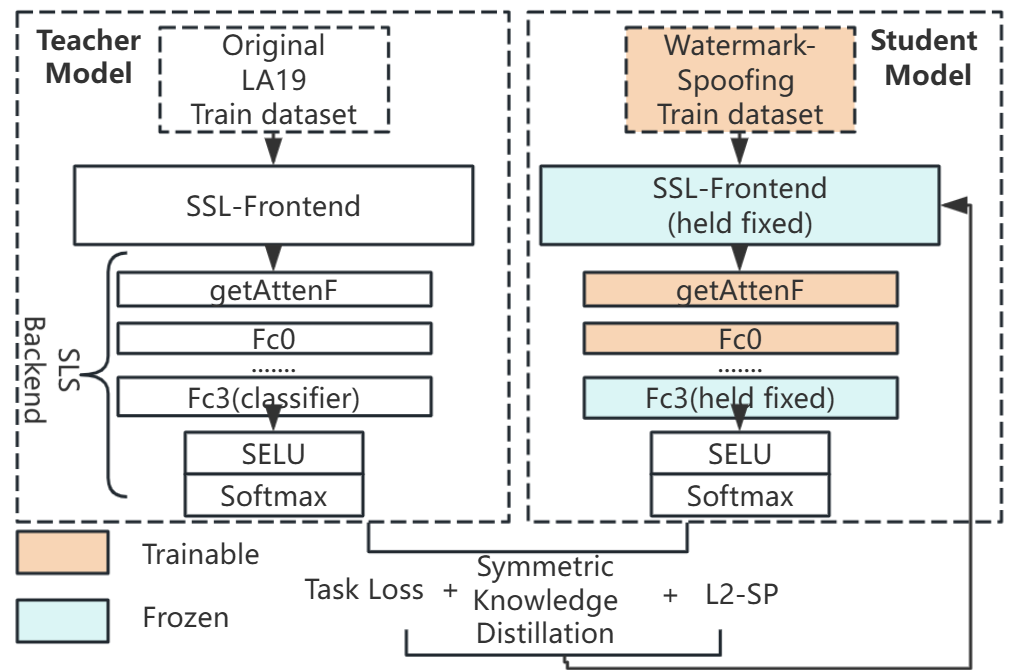}
    \caption{The Knowledge-Preserving Watermark Learning framework: the model is first trained normally as\cite{zhang2024xlsrsls}, then adapted with the SSL front end and classifier frozen while only intermediate layers are updated}
    \label{fig:placeholder}
\end{figure}

To reduce the negative impact of watermarks while preserving the model’s spoofing detection ability on original data, we propose \textit{Knowledge-Preserving Watermark Learning (KPWL)} as shown in Fig. \ref{fig:placeholder}.

\textbf{Original Baseline Pretraining:}
We first train the antispoofing model on the original LA19 data using a standard supervised objective, establishing a strong original-domain baseline, in this case, the XLSR+SLS model.

\textbf{Knowledge-Preserving Watermark Learning:}
Next, the \textit{Watermark-Spoofing Training dataset} is used for model adaptation. In this phase, the SSL front end (e.g., XLSR) and the final classification layer (e.g., \texttt{fc3} in SLS) are kept frozen (non-trainable), while only the intermediate backend layers are updated.
Training is conducted with a slightly higher learning rate than in previous phase and for 2 epochs, enabling rapid adaptation with minimal drift.
To stabilize this process, a teacher–student framework is employed, incorporating symmetric knowledge distillation and parameter anchoring (L2-SP), as shown in Eq. \ref{eq:totall}.
\begin{equation}
\label{eq:totall}
\begin{aligned}
\mathcal{L}
\,&=\,
\underbrace{\mathcal{L}_{\mathrm{task}}}_{\text{supervised}}
\;+\;
\beta\,\underbrace{\mathcal{L}_{\mathrm{KD}}}_{\text{teacher--student}}
\;+\;
\mu\,\underbrace{\mathcal{L}_{\mathrm{L2\text{-}SP}}}_{\text{parameter anchor}},
\\
&\mathcal{L}_{\mathrm{KD}}
= D_{\mathrm{KL}}\!\big(p_T \,\|\, p_S\big)
 \;+\; D_{\mathrm{KL}}\!\big(p_S \,\|\, p_T\big),
\\
&\mathcal{L}_{\mathrm{L2\text{-}SP}}
= \sum_{i\in\mathcal{I}_{\mathrm{trainable}}} \big\| w_i - w_i^{(0)} \big\|_2^2,
\end{aligned}
\end{equation}
where \(\mathcal{L}_{\mathrm{task}}\) is a weighted cross-entropy loss to optimize classification performance. \(\mathcal{L}_{\mathrm{KD}}\) is a symmetric knowledge distillation loss that constrains the adapted model’s predictions to remain close to those of the frozen teacher snapshot. In addition, L2-SP regularization \(\mathcal{L}_{\mathrm{L2\text{-}SP}}\) penalizes deviation of trainable parameters from their initialization at the start of Knowledge-Preserving Watermark Learning phase, thereby reducing parameter drift during adaptation.


Freezing the SSL front end preserves low-level features against watermark-induced shifts, while fixing the classifier maintains the decision boundary. Symmetric Knowledge Distillation constrains the \emph{outputs} to remain close to the pre-adaptation model, while L2-SP constrains the \emph{trainable parameters} not to drift far from the Knowledge-Preserving Watermark Learning phase starting point. Together, these choices allow the model to maintain its original-domain performance while reliably adapting to watermarked inputs.

 \section{Experimental Analysis}

\subsection{Experimental Settings}
\noindent\textbf{Preprocessing:}
All audio is resampled to 16\,kHz and normalized to 64{,}600 samples by tiling or truncation. 
RawBoost~\cite{tak2022rawboost} is applied with colored noise at random SNRs, and the resulting waveforms are converted to single-channel format.

\noindent\textbf{Training:}
Original Baseline Pretraining phase trains on the LA19 split using Adam with a learning rate of \(1\times10^{-7}\), and weight decay of \(10^{-4}\), optimized with class-weighted cross-entropy for up to 50 epochs with early stopping. 
 continues from Original Baseline Pretraining phase, trained with\textit{Watermark-Spoofing Training dataset}, the watermarked version of LA19, using Adam with a learning rate of \(5\times10^{-7}\), and weight decay of \(10^{-4}\). 
The SSL frontend and classifier are frozen while only intermediate layers are updated, trained for 2 epochs using class-weighted NLL with symmetric knowledge distillation (\(\beta=0.3\)) and L2-SP regularization (\(\mu=10^{-4}\)).

\noindent\textbf{Evaluation:}
To assess the effectiveness of the KPWL, three variations of XLSR+SLS were compared: 
(i) Model trained solely on original LA19 data (hereafter referred to as the baseline model), 
(ii) Model trained on \textit{Watermark-Spoofing Training dataset}. (hereafter referred to as the watermarked model), and 
(iii) Model trained using the KPWL method (hereafter referred to as the KPWL model). 
These models were evaluated on the \textit{LA21}, \textit{DF21}, and \textit{In-the-Wild} datasets, each tested in their original form and in watermarked versions—collectively referred to as the \textit{Watermark-Spoofing Seen Evaluation dataset}—as well as on a more challenging \textit{Watermark-Spoofing Unseen Evaluation dataset}, which was generated using the same procedure as the seen set but with a distinct set of watermarking methods~\cite{saadi2019normspace,nugraha2011dsss,adhiyaksa2022reversible,alsabhany2020digital,PAVLOVIC2022103381,singh24_interspeech}.

\subsection{Experimental Results}

As shown in Table \ref{tab:xlsr_sls_three}, the proposed KPWL model consistently outperforms the baseline on the \textit{Watermark-Spoofing Seen Evaluation dataset}, reducing the EER on ITW with 75\% watermarked data from 8.46\% to 7.92\%. It also maintains comparable performance on clean data (3.06\% vs. 3.02\% on LA21). In contrast, the watermarked model gains robustness against domain shift caused by domain shifts as watermark density increases but sacrifices clean-domain accuracy. These results show that KPWL suppresses watermark-induced domain shifts while preserving spoofing detection capability, indicating that existing models can adapt to watermarked data when guided by an appropriate training strategy.

\begin{table}[t]
\centering
\small
\setlength{\tabcolsep}{4pt}
\caption{Comparison of three XLSR+SLS variants on different test datasets under \textit{Watermark-Spoofing Seen Evaluation dataset}. $\Delta$ denotes relative increase from 0\%.}
\label{tab:xlsr_sls_three}
\begin{tabular}{l l | cccc }
\toprule
\textbf{Dataset} & \textbf{Model} & 75\% & 50\% & 25\% & 0\% \\
\midrule
\multirow{3}{*}{LA21}
 & Baseline           & 3.68 & 3.52 & 3.35 & 3.02  \\
 & Watermarked  & 3.28 & 3.25 & 3.23 & 3.17  \\
 & KPWL           & 3.21 & 3.18 & 3.12 & 3.06  \\
\midrule
\multirow{3}{*}{ITW}
 & Baseline           & 8.46 & 7.83 & 7.57 & 7.32 \\
 & Watermarked  & 9.03 & 8.75 & 8.57 & 8.21  \\
 & KPWL           & 7.92 & 7.74 & 7.60 & 7.37  \\
\midrule
\multirow{3}{*}{DF21}
 & Baseline           & 2.23 & 2.13 & 2.17 & 2.01  \\
 & Watermarked & 2.13 & 2.07 & 2.00 & 1.87 \\
 & KPWL            & 2.04 & 1.95 & 1.92 & 1.74 \\
\bottomrule
\end{tabular}
\end{table}

As shown in Table \ref{tab:watermark_models}, performance changes were far more drastic on the \textit{Watermark-Spoofing Unseen Evaluation dataset}. The baseline model outperformed both the watermarked and KPWL variants (9.94\% vs. 10.21\% and 11.22\% on 75\%-watermarked data from the \textit{Watermark-Spoofing Seen Evaluation dataset} LA21), indicating that the latter two may have over-adapted to the specific watermarking methods used to generate the \textit{Watermark-Spoofing Training dataset}. This suggests that current anti-spoofing models lack robustness to diverse distortions, with watermarking exposing a critical gap that warrants further investigation.


\label{sec:print}
\begin{table}[t]
\centering
\small
\setlength{\tabcolsep}{4pt}
\caption{EER(\%) of different variants of XLSR+SLS under \textit{Watermark-Spoofing Unseen Evaluation dataset} across different test datasets.}
\label{tab:watermark_models}
\begin{tabular}{l l | cccc}
\toprule
Dataset & Model        & 75\%  & 50\%  & 25\%  & 0\%  \\
\midrule
\multirow{3}{*}{LA21}
 
 & Baseline     &  9.94 & 8.01 & 5.72 & 3.07 \\
 & Watermarked  & 10.21 & 8.00 & 5.64 & 3.17 \\
 & KPWL    & 11.22 & 8.47 & 5.66 & 3.04 \\
\midrule
\multirow{3}{*}{ITW}
 
 & Baseline     & 13.41 & 11.43 & 9.36 & 7.32 \\
 & Watermarked  & 15.66 & 13.15 & 10.68 & 8.62 \\
 & KPWL     & 14.78 & 12.27 & 9.71 & 7.37 \\
\midrule
\multirow{3}{*}{DF21}
 
 & Baseline     &  8.42 & 6.34 & 4.31 & 2.01 \\
 & Watermarked  &  8.60 & 6.40 & 4.20 & 1.87 \\
 & KPWL     & 10.05 & 7.24 & 4.50 & 1.75 \\
\bottomrule
\end{tabular}
\end{table}
\section{Conclusion}
This work presents the first study on how audio watermarking impacts anti-spoofing detection systems. Experiments across diverse datasets and models show that watermarking can significantly degrade detector performance during both training and evaluation. To support future research, we constructed watermark-augmented training and evaluation datasets. To mitigate mentioned issues, we propose KPWL, a teacher–student adaptation framework that preserves original-domain decision boundaries while adapting to watermark-induced shifts. These findings highlight audio watermarking as an overlooked source of domain shift and underscore the need for anti-spoofing models robust to real-world watermark contamination, while handling unseen watermark scenarios remains an open challenge.

\vfill\pagebreak
\small

\bibliographystyle{IEEEbib}
\bibliography{strings,refs}

\end{document}